\documentclass[letterpaper]{article} 
\usepackage{aaai25}  
\usepackage{times}  
\usepackage{helvet}  
\usepackage{courier}  
\usepackage[hyphens]{url}  
\usepackage{graphicx} 
\usepackage{natbib}  
\usepackage{caption} 
\usepackage{amssymb}
\usepackage{amsmath}
\usepackage{algorithm}
\usepackage{algorithmic}
\usepackage{subcaption}
\usepackage{makecell}
\usepackage{multirow}
\usepackage{color}
\usepackage[table]{xcolor}
\usepackage{colortbl}
\usepackage{xcolor}
\usepackage{xpatch}
\usepackage{booktabs}
\usepackage{newfloat}
\usepackage{listings}
\DeclareCaptionStyle{ruled}{labelfont=normalfont,labelsep=colon,strut=off} 
\floatstyle{ruled}
\newfloat{listing}{tb}{lst}{}
\floatname{listing}{Listing}
\title{CDW-CoT: Clustered Distance-Weighted Chain-of-Thoughts Reasoning}
\author{
    Yuanheng Fang\textsuperscript{\rm 1},
    Guoqing Chao\thanks{Corresponding author}\textsuperscript{\rm 1},
    Wenqiang Lei\textsuperscript{\rm 2},
    Shaobo Li\textsuperscript{\rm 1},
    Dianhui Chu\textsuperscript{\rm 1}
}
\affiliations {
    \textsuperscript{\rm 1}Harbin Institute of Technology, Weihai, 264209, Shandong, China\\
    \textsuperscript{\rm 2}Sichuan University, Chengdu, 610065, Sichuan, China\\
    23s130443@stu.hit.edu.cn, 
    guoqingchao@hit.edu.cn, 
    wenqianglei@scu.edu.cn, 
    lishaobo@hit.edu.cn, 
    chudh@hit.edu.cn
}

\usepackage{bibentry}

\begin{document}

\maketitle

\begin{abstract}

Large Language Models (LLMs) have recently achieved impressive results in complex reasoning tasks through Chain of Thought (CoT) prompting. However, most existing CoT methods rely on using the same prompts, whether manually designed or automatically generated, to handle the entire dataset. This one-size-fits-all approach may fail to meet the specific needs arising from the diversities within a single dataset. To solve this problem, we propose the Clustered Distance-Weighted Chain of Thought (CDW-CoT) method, which dynamically constructs prompts tailored to the characteristics of each data instance by integrating clustering and prompt optimization techniques. Our method employs clustering algorithms to categorize the dataset into distinct groups, from which a candidate pool of prompts is selected to reflect the inherent diversity within the dataset. For each cluster, CDW-CoT trains the optimal prompt probability distribution tailored to their specific characteristics. Finally, it dynamically constructs a unique prompt probability distribution for each test instance, based on its proximity to cluster centers, from which prompts are selected for reasoning. CDW-CoT consistently outperforms traditional CoT methods across six datasets, including commonsense, symbolic, and mathematical reasoning tasks. Specifically, when compared to manual CoT, CDW-CoT achieves an average accuracy improvement of 25.34\% on LLaMA2 (13B) and 15.72\% on LLaMA3 (8B).

\end{abstract}
\section{Introduction}
Recent advancements in LLMs, such as GPT-3~\cite{gpt3}, LLama2~\cite{llama2}, and Llama3~\cite{llama3}, have significantly enhanced their capability to tackle complex reasoning tasks. Some studies~\cite{gpt3,lamda} have demonstrated LLMs' impressive performance in decomposing multi-step problems into manageable intermediate steps, resulting in more accurate and contextually relevant answers. A technique that has gained prominence in this context is CoT prompting, which systematically structures the reasoning process into a series of intermediate steps. This method has been shown to significantly improve the model's performance on complex tasks across various domains~\cite{Elicits}.

Initially, CoT prompting involved embedding manually crafted exemplars within the model's prompt to guide its reasoning process—a method that was effective but labor-intensive and not scalable~\cite{Elicits}. This approach evolved into Zero-Shot-CoT~\cite{zero_shot}, which allowed models to engage in reasoning without task-specific exemplars, relying on generic prompts to elicit intermediate reasoning steps. However, the absence of tailored guidance often limited its efficacy in more complex or domain-specific tasks.
\begin{figure}[htb]
    \centering
    \includegraphics[width=\columnwidth]{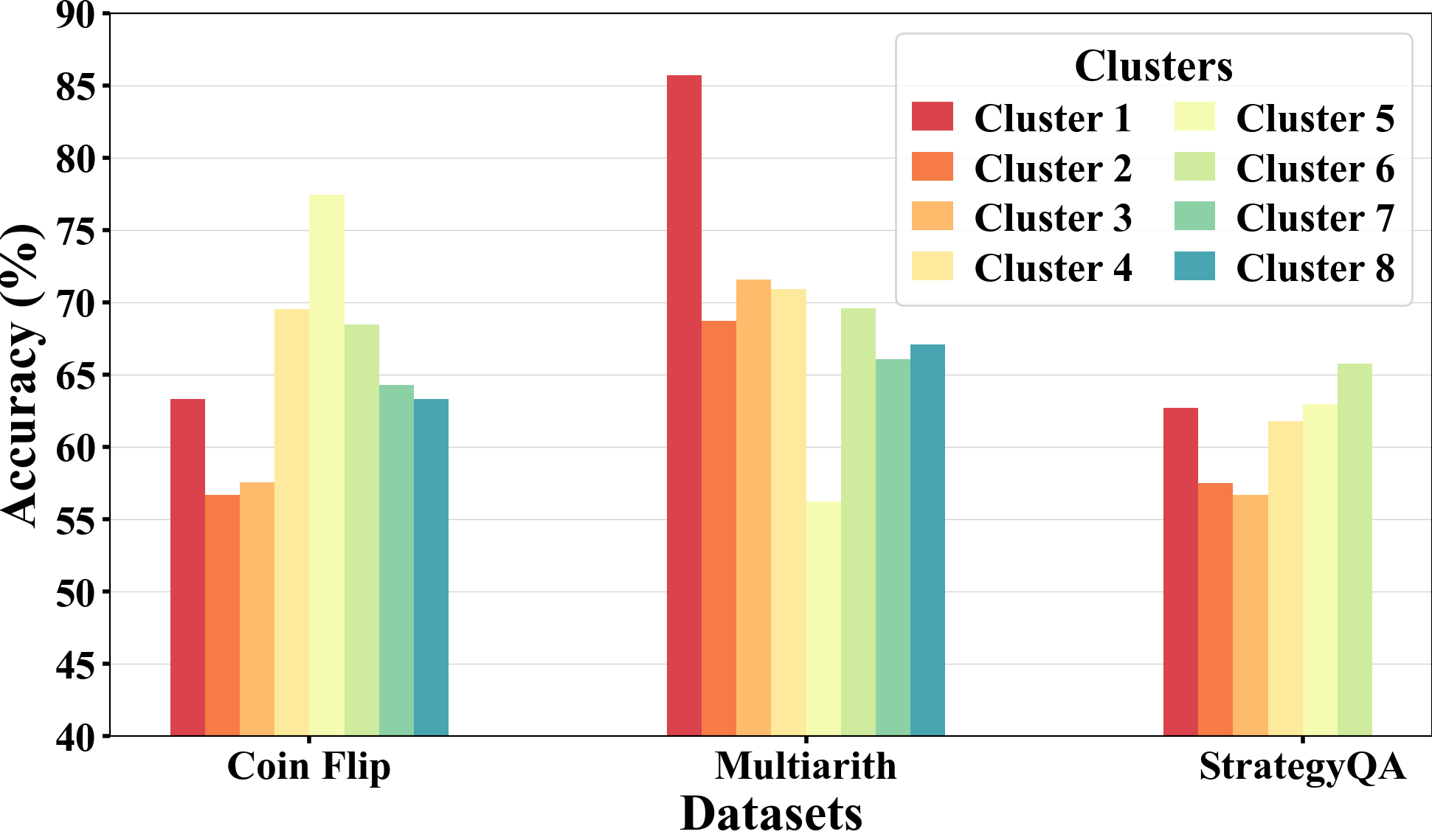} 
    \caption{Using the same prompts for all instances in the dataset resulted in significant performance variability across different clusters, highlighting the limitations of Auto-CoT in addressing diverse reasoning demands within different data categories. This underscores the need for tailored prompt strategies.}

    \label{fig:reduced-dataset-performance}
\end{figure}

To address the intensive manual efforts required by Manual-CoT, the Auto-CoT paradigm has been proposed~\cite{automatic}. This method automates the generation of reasoning chains by clustering related questions and selecting a representative question from each cluster to generate a reasoning chain using simple heuristics. Recent works \cite{chu2023survey} focus on further automating~\cite{automatic_pool} and refining the CoT generation process. Techniques such as Enhanced Reasoning~\cite{CoK,mot,wu2023chain}, Voting and Ranking~\cite{complexity,step_aware_verifier,li2022advance}, and Verification and Refinement~\cite{Faithful,synthetic,adaptive_consistency,selfVerification,wang2022self} have been developed to enhance the quality and applicability of CoT across diverse tasks. Especially, Automate-CoT approach integrates variance-reduced policy gradient methods to optimize the selection of CoT exemplars, significantly reducing the dependency on manual prompt engineering.

Despite these advancements, Automatic Chain of Thought Prompting still encounters significant challenges, particularly because most of them use the same prompts for all instances in the dataset. As exemplified by Auto-CoT \cite{automatic} shown in Figure~\ref{fig:reduced-dataset-performance}, this approach results in considerable performance variability across different clusters. This variability highlights its inability to effectively address the diverse reasoning demands of different data categories, emphasizing the necessity for more adaptive techniques that can tailor prompts to the unique characteristics of each cluster. 

To address the limitations inherent in manual and uniform prompt strategies across diverse datasets, we introduce the CDW-CoT framework. This method innovatively combines clustering with dynamic prompt optimization to enhance the adaptability and precision in reasoning tasks. By segmenting the dataset into distinct clusters, CDW-CoT harnesses the unique characteristics of each group to generate a tailored prompt candidate pool. For each cluster, we calculate an optimal prompt probability distribution, finely tuned to the specific demands and nuances of the data. Additionally, our framework incorporates a distance-weighted prompt selection mechanism that dynamically adapts reasoning strategies based on the proximity of test instances to the cluster centers. This ensures that each reasoning step is contextually informed and effectively customized, significantly improving the reasoning accuracy. Experiment results on six datasets show the superiority of our proposed method CDW-CoT over the state-of-the-art methods.

The main contributions of our work are summarized as follows:
\begin{itemize}
    \item We leverage the clustering technique to produce a diverse prompt candidate pool that mining the category-specific information sufficiently, enhancing the relevance and effectiveness of prompts for different clusters within the same dataset.
    \item Our framework calculates the optimal prompt probability distributions for each cluster within the dataset, effectively treating datasets as distinct clusters and enabling highly targeted reasoning approaches tailored to the unique characteristics of each group.
    \item We introduce a method for employing distance-weighted calculations for each test instance's prompt probability distribution, which refines and tailors the reasoning process of large language models to the specific requirements of each instance.
    \item Our empirical evaluations confirm that the CDW-CoT framework substantially outperforms traditional CoT methods, achieving the state-of-the-art accuracy across multiple datasets.

\end{itemize}
\section{Related Works}
\subsection{Chain of Thought Prompting}
CoT Prompting enhances logical reasoning in LLMs like GPT-3, developed as the models increased in scale~\cite{gpt3}. Wei et al. first introduced CoT, using manually constructed detailed prompts to systematically guide LLMs through each logical step, significantly enhancing reasoning transparency and accuracy~\cite{Elicits}. Building on this foundational work, Zero-Shot-CoT employs the simple prompt ``Let’s think step by step" to facilitate unsupervised reasoning, effectively enabling CoT without predefined examples~\cite{zero_shot}.
\subsection{Automatic Chain of Thought Prompting}
Addressing the accuracy challenges in Zero-Shot-CoT and the resource intensity of Few-Shot CoT, Auto-CoT automates reasoning chain generation. This method clusters related questions, using cluster centers as prompts, thereby reducing manual labor and improving scalability~\cite{automatic}. Building on this, complexity-based prompting selects prompts based on their reasoning complexity, which has been shown to improve performance on multi-step reasoning tasks significantly~\cite{complexity}. Furthermore, self-verification techniques introduced in studies allow models to cross-check and refine their outputs~\cite{selfVerification}. In the context of mathematical reasoning, the MathPrompter framework validates results by leveraging different algebraic expressions or Python functions to solve problems~\cite{mathprompter}.
\subsection{Policy Gradient Optimization Methods}
The Black-Box Discrete Prompt Learning (BDPL) employs variance-reduced policy gradients to optimize prompts efficiently, enhancing LLM performance without direct access to model parameters~\cite{BBPL}. Following this, the Black-Box Prompt Optimization (BPO) further refines these advancements by aligning LLM outputs with user preferences through optimized prompts, improving user interactions and satisfaction~\cite{BPO}. Dynamic Prompt Learning via Policy Gradient (PROMPTPG) further refines this approach by dynamically selecting in-context examples that optimize reasoning tasks, particularly in complex domains like mathematics~\cite{DynamicPG}. Building on these strategies, the Automatic Prompt Augmentation and Selection method extends the application of policy gradient methods to CoT prompting, automating both the generation and the optimization of reasoning chains~\cite{automatic_pool}.
\section{CDW-CoT Model}
\begin{figure*}[ht]
\centering
\includegraphics[width=\textwidth]{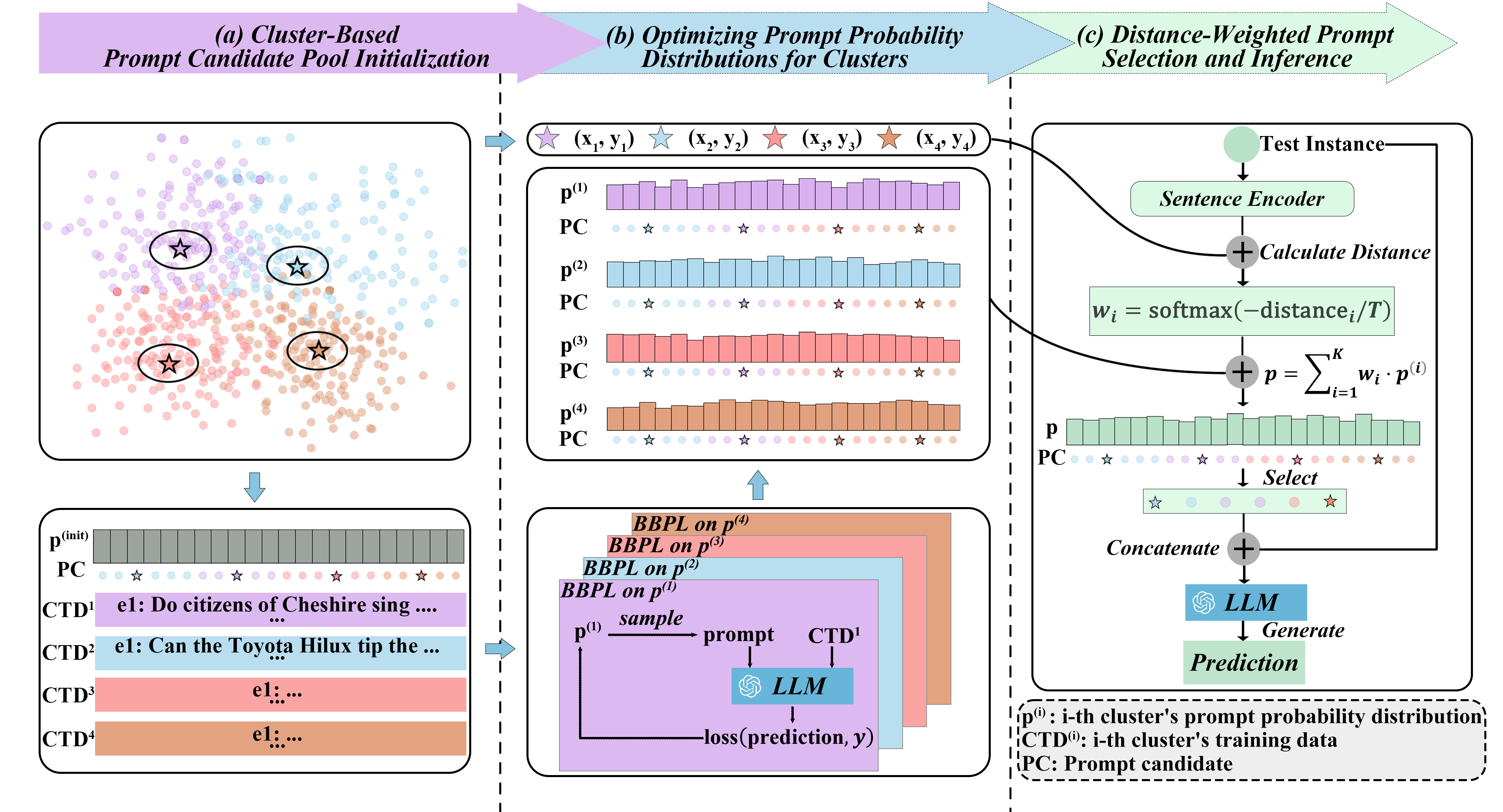}
\caption{Framework of the proposed CDW-CoT. (a) After clustering, prompt candidates are generated based on the cluster centers. \( \text{CTD}^{(i)} \) and cluster center coordinates \((x_i, y_i)\) are also obtained. (b) For each cluster, \( p^{(i)} \) is initially set to \( p^{(\text{init})} \) and then optimized through Black-Box Prompt Learning (BBPL) to achieve the optimal distribution. (c) For test instance, a distance-weighted prompt probability distribution is constructed to select prompts and perform reasoning.
}

\label{fig:model}
\end{figure*}
\label{intro}
In this section, we introduce our proposed CDW-CoT model, as depicted in Figure~\ref{fig:model}. The CDW-CoT is composed of the three components: cluster-based prompt candidate pool initialization, optimizing prompt probability distributions for clusters and distance-weighted prompt selection and inference.

\subsection{Cluster-Based Prompt Candidate Pool Initialization}

The dataset \(D = \{(x_i, y_i)\}_{i=1}^N\) consists of \(N\) question-answer pairs. Each instance \(x_i\) is transformed into the vector embeddings \(\{e_i\}\) using a pre-trained sentence transformer and clustered into \(K\) groups via K-means.

As illustrated in Figure \ref{fig:model}(a), following the clustering process, the preliminary selection of prompt candidates begins from the centroid of each cluster. The number of candidates selected from each cluster \(S_c\) is based on the proportion of that cluster's data within the overall dataset. Once the preliminary prompt candidate pool is established, it is refined into the final prompt candidates (PC) through zero-shot-CoT \cite{zero_shot}. 
The entire process is detailed in Algorithm \ref{alg:prompt_generation}.

Simultaneously, we establish an initial prompt probability distribution \( p^{(\text{init})} \), where each candidate in the pool is assigned an equal probability. This balanced distribution, along with the cluster-specific training data \( \text{CTD}^{(i)} \), serves as the foundation for the next phase of the model training process.

Furthermore, the coordinates of each cluster's centroid, obtained during the clustering process, are stored to use for calculating the distance of the test instance to them. These coordinates play a critical role in  the model's final phase: distance-weighted prompt selection and inference, where they guide the distance-weighted prompt probability distribution and ensure the model adapts effectively to new, unseen instances.

\begin{algorithm}[H]
\caption{Cluster-Based Prompt Candidate Pool Initialization}
\label{alg:prompt_generation}
\textbf{Input}: \(X = \{x_1, \dots, x_N\}\), \(Y = \{y_1, \dots, y_N\}\), Number of Clusters \(K\), Pool Size \(S\) \\
\textbf{Output}: Prompt Candidates \(PC\)
\begin{algorithmic}[1]
\STATE \( \{e_i\} \leftarrow \text{SentenceTransformer}(X) \)  
\STATE \(\text{cluster\_assignment} \leftarrow \text{K-Means}(K, \{e_i\})\)  

\STATE Initialize cluster data structure \(C = [\,]\)  
\FOR{\(i \leftarrow 1\) \textbf{to} \(N\)}
    \STATE \(c_i \leftarrow \text{cluster\_assignment}[i]\)  
    \STATE \(d \leftarrow \text{EuclideanDistance}(e_i, \text{cluster\_centers}[c_i])\)  
    \STATE Add \( (x_i, y_i, d) \) to \( C[c_i] \)
\ENDFOR

\STATE Prepare to build prompt candidate pool:
\FOR{\(c \leftarrow 1\) \textbf{to} \(K\)}
    \STATE Compute \(S_c \leftarrow \left\lfloor \frac{|C[c]|}{N} \times S \right\rfloor \)  
    \STATE Sort \(C[c]\) by distance \(d\)  
    \STATE \(P_c \leftarrow \text{SelectTop}(C[c], \text{size} = S_c)\)  
    \STATE Add \( P_c \) to \( P \) 
\ENDFOR

\STATE \(PC \leftarrow \text{ZeroShotCoT}(P, \text{LLM})\)  
\STATE \textbf{return} \(PC\)  
\end{algorithmic}
\end{algorithm}

\subsection{Optimizing Prompt Probability Distributions for Clusters}

As illustrated in Figure \ref{fig:model}(b), the optimization of prompt probability distribution for each cluster is conducted using the BBPL method. The process begins by setting the initial distribution \( p^{(i)} \) for each cluster to a uniform distribution \( p^{(\text{init})} \). This distribution is then refined through gradient descent, based on feedback from the training process with \( \text{CTD}^{(i)} \).

For each cluster \( i \), prompts are sampled according to the \( p^{(i)} \). These prompts, along with the \( \text{CTD}^{(i)} \), are input into the LLM, which returns the prediction and computes the corresponding loss.

The gradient for each prompt is computed as:

\begin{equation}
\delta = -\frac{1}{p^{(i)}},
\end{equation}
where \( p^{(i)} \) represents the prompt probability matrix for cluster \( i \).

These gradients are adjusted based on the actual usage of prompts during training:

\begin{equation}
\delta_{k,m,n} = \begin{cases} 
-\delta_{k,m,n} & \text{if } n \text{ is selected} \\
\delta_{k,m,n} & \text{otherwise.}
\end{cases}
\end{equation}

Here, \( k \) indexes the sample within the batch, \( m \) represents the prompt, and \( n \) corresponds to the indices of the prompts sampled. Adjustments are weighted by the deviation of each sample's loss from the batch average:

\begin{equation}
\text{Gradient} = \sum_{k=1}^{\text{s}} \frac{L_k - L_{\text{avg}}}{\text{s} - 1} \times \delta_k,
\end{equation}
where \( s \) is the sample size used in the optimization process. Using the aggregated gradient and learning rate \(\eta\), the probability matrix \( p^{(i)} \) is updated according to the following formula:

\begin{equation}
p^{(i)}_{mn} \leftarrow p^{(i)}_{mn} - \eta \cdot \text{Gradient}_{mn}.
\end{equation}

Probabilities are then normalized and clipped within the range [0,1] to ensure stability:

\begin{equation}
p^{(i)}_{mn} \leftarrow \max(\min(p^{(i)}_{mn}, 1), 0).
\end{equation}

The optimized prompt probabilities are validated on a validation dataset. If the performance improves, these updated settings are used for future operations. 

Through this process, we obtain the optimal prompt probability distribution for each cluster, as depicted in Figure \ref{fig:model}(b).

\subsection{Distance-Weighted Prompt Selection and Inference}

This subsection describes how we construct unique prompt probability distribution for each test instance through distance weighting, using the optimal prompt probability distribution for each cluster, and the coordinates of cluster centers. The obtained prompt probability distribution is used to select the final prompts that are concatenated with test instance and input into the LLM, as illustrated in Figure \ref{fig:model}(c).

\subsubsection{Distance Calculation}
For each test instance, its embedding obtained through a sentence Transformer is compared with the cluster centers to compute the Euclidean distances. These distances reflect the instance's similarity to each cluster.

\subsubsection{Weight Calculation}
The computed distances are converted into weights using a temperature-scaled softmax function:

\begin{equation}
\text{weights} = \frac{\exp(-\text{distances} / T)}{\sum \exp(-\text{distances} / T)},
\end{equation}
where \( T \) is a temperature parameter that controls the sensitivity to distance variations.

\subsubsection{Prompt Distribution Calculation}
The prompt probability distribution for the test instance is calculated by weighting the optimal prompt distributions of each cluster:

\begin{equation}
\text{p} = \sum_{i=1}^{K} \text{weights}_i \cdot p^{(i)},
\end{equation}
where \( K \) is the number of clusters. This weighted combination tailors the prompt probability distribution to the specific characteristics of the test instance.

\subsubsection{Query Execution and Evaluation}
Prompts are selected based on the computed distribution and then concatenated with the original test instance. The LLM uses this concatenated input to generate a response, which is subsequently evaluated against the actual answer to assess the accuracy and effectiveness.

The steps of this process are outlined in Algorithm \ref{alg:testing_optimized_prompts}, demonstrating the implementation of distance-weighted prompt probability distribution and its impact on inference.




\begin{algorithm}[H]
\caption{Distance-Weighted Prompt Selection and Inference}
\label{alg:testing_optimized_prompts}
\textbf{Input}: Test dataset \(D_{\text{test}}\), Cluster centers \(C\), Temperature \(T\), Optimized prompt probabilities for each cluster \(p^{(i)}\) \\
\textbf{Output}: Evaluated responses \(R\)
\begin{algorithmic}[1] 
\STATE Initialize \( \text{weights} \) as an empty list
\STATE Initialize \(R\) as an empty list to store responses
\FOR{each instance \(q\) in \(D_{\text{test}}\)}
    \STATE \(e_q \leftarrow \text{SentenceTransformer}(q)\)
    \FOR{each center \(c\) in \(C\)}
        \STATE Compute distance: \( \text{distance} = \text{Euclidean}(e_q, c) \)
        \STATE Compute weight: \( \text{weight} = \exp(-\text{distance} / T) \)
        \STATE Append \( \text{weight} \) to \( \text{weights} \)
    \ENDFOR
    \STATE Normalize weights: \( \text{weights} \leftarrow \frac{\text{weights}}{\sum \text{weights}} \)
    \STATE Compute the prompt probability distribution for q: \( \text{p} = \sum_{i=1}^{K} \text{weights}_i \cdot p^{(i)} \)
    \STATE Select prompt using \( \text{p} \)
    \STATE Input prompt and \(q\) into LLM to generate response
    \STATE Evaluate response accuracy and append  to \(R\)
\ENDFOR
\STATE \textbf{return} \(R\)
\end{algorithmic}
\end{algorithm}
\begin{table*}[ht]
    \centering
    \begin{tabular}{lcccccc}
        \toprule
        \multirow{2}{*}{Method} & \multicolumn{2}{c}{Commonsense Reasoning} & \multicolumn{2}{c}{Symbolic Reasoning} & \multicolumn{2}{c}{Mathematical Reasoning} \\
        \cmidrule(lr){2-3} \cmidrule(lr){4-5} \cmidrule(lr){6-7}
        & CSQA & StrategyQA & Letter & Coin & MultiArith & AQuA \\
        \midrule
        \multicolumn{7}{c}{\textbf{LLaMA2 (13B)}} \\
        Zero-Shot-CoT~\cite{zero_shot} & 32.68 & 48.41 & 30.20 & 51.80 & 71.00 & 30.31 \\
        Auto-CoT~\cite{automatic} & 51.09 & 56.24 & 30.80 & 51.00 & 44.17 & 24.02 \\
        Manual-CoT~\cite{Elicits} & 46.52 & 60.48 & 15.80 & 47.60 & 44.17 & 30.31 \\
        \textbf{CDW-CoT (ours)} & \textbf{61.41} & \textbf{70.06} & \textbf{82.67} & \textbf{61.33} & \textbf{85.56} & \textbf{35.89} \\
        \midrule
        \multicolumn{7}{c}{\textbf{LLaMA3 (8B)}} \\
        Zero-Shot-CoT~\cite{zero_shot} & 60.52 & 66.72 & 76.67 & 44.40 & 90.00 & 48.03 \\
        Auto-CoT~\cite{automatic} & 69.57 & 60.57 & 50.60 & 60.80 & 68.83 & 31.10 \\
        Manual-CoT~\cite{Elicits} & 56.84 & 57.51 & 61.40 & 60.20 & 85.17 & 32.68 \\
        \textbf{CDW-CoT (ours)} & \textbf{72.15} & \textbf{67.44} & \textbf{82.67} & \textbf{70.70} & \textbf{95.17} & \textbf{58.97} \\
        \bottomrule
    \end{tabular}
    \caption{Comparative exact match accuracy across various datasets using LLaMA2 (13B) and LLaMA3 (8B) models. The CDW-CoT method consistently outperforms traditional CoT methods in all tested reasoning tasks and datasets, improving accuracy for both models. }
    \label{tab:Experiment_result}
\end{table*}

\section{Experiments and Results}
\label{experiments}

\subsection{Experiments Setup}
\subsubsection{Tasks and Datasets}
We evaluated the CoT frameworks on six datasets across three categories of reasoning tasks, which are listed as follows:

\begin{itemize}
    \item \textbf{Commonsense Reasoning}:
    
\textbf{CommonsenseQA (CSQA)}~\cite{CSQA}: A widely used dataset for evaluating commonsense reasoning through multiple-choice questions that require inferencing based on prior knowledge and context.

\textbf{StrategyQA}~\cite{strategyQA}: It contains questions requiring implicit multi-hop reasoning to derive yes/no answers, testing the model's ability to connect various pieces of information logically.
    \item \textbf{Symbolic Reasoning}:
    
\textbf{Letter}~\cite{Elicits}: It involves tasks like last letter concatenation, designed to test the symbolic reasoning capabilities of models.
    
\textbf{Coin}~\cite{Elicits}: It focuses on determining the state of a coin after a series of flips, evaluating the model’s ability to track state changes through symbolic manipulations.
    
    \item \textbf{Mathematical Reasoning}:
    
\textbf{MultiArith}~\cite{MultiArith}: It consists of multi-step arithmetic word problems that require a sequence of operations to reach the solution, testing multi-step reasoning in arithmetic contexts.
    
\textbf{AQuA}~\cite{Aqua}: It includes complex arithmetic word problems with multiple-choice answers, providing a benchmark for evaluating sophisticated reasoning and calculation skills.
\end{itemize}


\subsubsection{Models and Baselines}
We conducted comparative experiments using both the LLaMA2 (13B) and LLaMA3 (8B) models, running on two NVIDIA 4090 GPUs locally. The LLaMA2 (13B) model was selected for its easy-use, while LLaMA3 (8B) was chosen to evaluate the scalability of our approach across different large language models.
To evaluate the performance of our CDW-CoT framework, we compared it against several baseline methods implemented on the same LLM:
\begin{itemize}
\item \textbf{Zero-Shot-CoT}~\cite{zero_shot}: It uses a simple prompt like ``Let's think step by step" without requiring prior demonstrations.

\item \textbf{Auto-CoT}~\cite{automatic}: It automates reasoning chain generation by clustering similar questions and and using the cluster centers as prompts.

\item \textbf{Manual-CoT}~\cite{Elicits}: It involves crafting manually designed reasoning chains, tailored with specific demonstrations for each dataset.
\end{itemize}

To show the superiority of our method, the number of prompts used in our work was kept less than or equal to those used in other methods, since more prompts typically yield better performance. The num of prompts used in the CDW-CoT framework were: 6 (CommonsenseQA), 5 (StrategyQA), 4 (Letter), 6 (Coin), 5 (MultiArith), and 4 (AQuA).

\subsubsection{Data Split and Number of Clusters Identification}

Datasets were divided into training, evaluation, and test subsets with proportions of approximately 60\%, 25\%, and 15\%, respectively \cite{test_split}. After dividing the data, we identified the number of clusters according to the Auto-CoT setup, and then adjusted the number of clusters for certain datasets from the default 8 to 3, as shown in Table \ref{tab:data_clusters}.
\begin{table}[ht]
\centering
\begin{tabular}{l|cccc|c}
\toprule
\textbf{Dataset} & \textbf{Total} & \textbf{Train} & \textbf{Eval} & \textbf{Test} & \textbf{\#Clusters} \\
\midrule
CSQA & 1,221 & 725 & 312 & 184 & 7 \\
StrategyQA    & 2,290 & 1,362 & 584 & 344 & 6 \\
Letter        & 500 & 297 & 128 & 75 & 4 \\
Coin          & 500 & 297 & 128 & 75 & 3 \\
MultiArith    & 600 & 357 & 153 & 90 & 3 \\
AQuA          & 254 & 150 & 65 & 39 & 4 \\
\bottomrule
\end{tabular}
\caption{Data Split and Number of Clusters Statistics.}
\label{tab:data_clusters}
\end{table}

\paragraph{Prompt Engineering}
Configuring prompts effectively is crucial for training models across diverse datasets. This phase involved three key parameters:
\begin{itemize}
\item \textbf{Pool Size}: We maintained a consistent pool of 40 potential prompts for each dataset to enable thorough exploration of diverse reasoning pathways.
\item \textbf{Sample Size}: During training, each instance was tested against five unique prompt combinations, assessing the effectiveness of various configurations.
\item \textbf{Temperature}: A temperature of 0.3 was used to optimize prompt selection during testing.
\end{itemize}
Our primary metric, exact match accuracy, measures the degree responses correctly answer the instances across various reasoning domains. As detailed in Table \ref{tab:Experiment_result}, our results demonstrate substantial performance improvements across all the tasks and both models used, underscoring the effectiveness of the CDW-CoT framework. For both the LLaMA2 (13B) and LLaMA3 (8B) models, we compared our method against the best baseline method among the three we evaluated.
\subsection{Main Results}
CDW-CoT consistently outperforms traditional CoT methods across various reasoning tasks and datasets, improving accuracy for both LLaMA2 and LLaMA3 models. The detailed results are as follows:

\textbf{Commonsense Reasoning:} For CommonsenseQA, CDW-CoT improved exact match accuracy by 10.32\% (51.09\% $\rightarrow$ 61.41\%) on LLaMA2 (13B) and by 2.58\% (69.57\% $\rightarrow$ 72.15\%) on LLaMA3 (8B). For StrategyQA, CDW-CoT increased accuracy by 9.58\% (60.48\% $\rightarrow$ 70.06\%) on LLaMA2 (13B) and by 0.72\% (66.72\% $\rightarrow$ 67.44\%) on LLaMA3 (8B).

\textbf{Symbolic Reasoning:} In the Letter dataset, CDW-CoT significantly improved accuracy by 51.87\% (30.80\% $\rightarrow$ 82.67\%) on LLaMA2 (13B) and by 6.07\% (76.67\% $\rightarrow$ 82.67\%) on LLaMA3 (8B). In the Coin dataset, CDW-CoT improved accuracy by 9.53\% (51.80\% $\rightarrow$ 61.33\%) on LLaMA2 (13B) and by 9.90\% (60.80\% $\rightarrow$ 70.70\%) on LLaMA3 (8B).

\textbf{Mathematical Reasoning:} CDW-CoT recorded a 14.56\% increase (71.00\% $\rightarrow$ 85.56\%) on MultiArith with LLaMA2 (13B) and a 5.17\% increase (90.00\% $\rightarrow$ 95.17\%) on LLaMA3 (8B). For AQuA, accuracy improved by 5.58\% (30.31\% $\rightarrow$ 35.89\%) on LLaMA2 (13B) and by 10.94\% (48.03\% $\rightarrow$ 58.97\%) on LLaMA3 (8B).

These results demonstrate that the CDW-CoT framework effectively enhances performance across a wide range of reasoning tasks, including commonsense reasoning, mathematical reasoning, and symbolic reasoning. The framework consistently outperforms Zero-Shot-CoT, Manual-CoT and even Auto-CoT and shows significant improvements across different LLMs, as confirmed by the results in Table \ref{tab:Experiment_result}.

\subsection{Ablation Study}

To evaluate the effectiveness of each component of our model, we conduct the experiments with different model versions by removing the corresponding component.

The three model versions are described as follows:

\begin{itemize}
    \item \textbf{Distance Weighting (Dist-W)}: This version implements the complete model, using clustering to generate optimal prompt probability distributions tailored to each category. It adjusts the reasoning process for each test instance by employing distance-weighted prompt distribution, enhancing specificity based on proximity to cluster centers.
    \item \textbf{Nearest Cluster (Near-C)}: This streamlined approach assigns the nearest cluster's prompt distribution to each test instance, omitting the computational complexity of distance weighting. This method emphasizes efficiency while still utilizing the benefits of clustering.
    \item \textbf{No Clustering (No-Clust)}: This baseline approach without clustering phase uses a single, global optimal prompt probability distribution, derived from the entire dataset and applied uniformly across all test instances.
\end{itemize}
The effectiveness of each model version was assessed using the same setups with the main experiments.

\begin{table}[ht]
\centering
\begin{tabular}{lccc}
\toprule
Dataset & Dist-W & Near-C & No-Clust  \\
\midrule
CSQA       & \textbf{61.41} & 53.26 & 60.33 \\
StrategyQA & \textbf{70.06} & 67.44 & 67.15 \\
Letter     & \textbf{82.67} & 81.33 & 81.11 \\
Coin       & \textbf{61.33} & 58.67 & 56.00 \\
MultiArith & \textbf{85.56} & 77.78 & 77.78 \\
AQuA       & \textbf{35.89} & 28.21 & 23.08 \\
\bottomrule
\end{tabular}
\caption{Ablation study of different model versions across datasets, showing percentage accuracies.}
\label{tab:ablation_study}
\end{table}


The results of our ablation study, as shown in Table \ref{tab:ablation_study}, clearly demonstrate the effectiveness of each component in the CDW-CoT framework. The important role of the Dist-W method is evident, as it consistently achieves the highest accuracy across all the datasets. This method highlights the importance of clustering and distance-based prompt optimization, allowing the model to adapt its reasoning pathways effectively by considering the unique aspects of each test instance. The Distance Weighting method is particularly successful in complex tasks such as MultiArith and Letter, where precise and context-aware reasoning is crucial.

The Near-C model, which only relies on the nearest cluster's prompt distribution without distance weighting, is limited in its capability to effectively use the optimal prompt probability distributions across multiple clusters. This constraint leads to  a 5.04\% decrease across datasets averagely, as shown in Table \ref{tab:ablation_study}.

The No-Clust model uses a uniform prompt distribution for all the instances, which reduces its effectiveness. Its lower performance in Table \ref{tab:ablation_study}, with an average decrease of approximately 5.24\% compared to the full model,  highlights the importance of constructing category-specific prompt distributions to address the distinct demands of various data categories effectively.

This ablation study confirms the robustness of our CDW-CoT framework, demonstrating that each component, particularly clustering and distance weighting, plays a crucial role in enhancing the reasoning performance.

\subsection{Sensitivity Analysis of Temperature}

Our analysis investigates the impact of the temperature parameter $T$ in our framework.

We explored the effects of temperature settings ranging from 0.1 to 1.0 measured with accuracy. The experiments were conducted using the LLaMA2(13B) model on StrategyQA and MultiArith.

\begin{figure}[ht]
    \centering
    \includegraphics[width=\columnwidth]{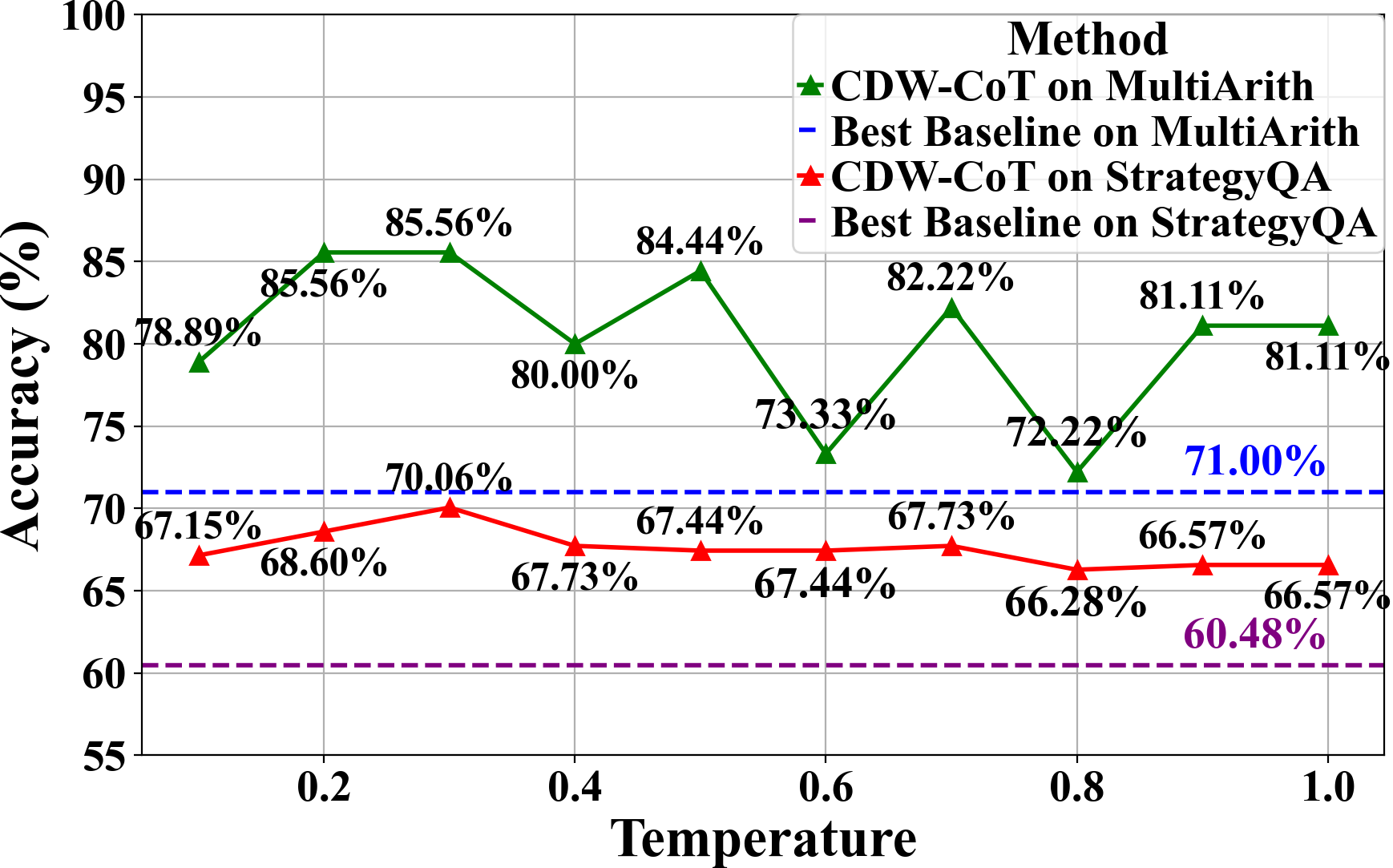}
    \caption{Sensitivity analysis of temperature for CDW-CoT on StrategyQA and MultiArith datasets.}
    \label{fig:temperature_effect}
\end{figure}

Temperature plays a pivotal role in the CDW-CoT framework, as evidenced by our detailed results depicted in Figure \ref{fig:temperature_effect}.  At a lower temperature of 0.1, the model becomes overly sensitive, disproportionately focusing on the nearest cluster even when it may not be the most relevant. This excessive sensitivity often leads to inaccuracies, especially when the query is ambiguously positioned relative to multiple clusters.

Conversely, at a temperature setting of 1.0, the model's performance declines due to an overly generalized approach that incorporates too much irrelevant cluster information. This almost uniform focus reduces the response accuracy and fails to fully leverage the optimal prompt distributions for each cluster.

Throughout all the temperatures, the CDW-CoT consistently surpasses the best baseline among the conventional methods compared, highlighting its superior reasoning capabilities and robust adaptability. The model achieves optimal performance at a temperature of 0.3, striking an effective balance between specificity and sensitivity. This setting allows the model to accurately concentrate on the most pertinent cluster features, thus maximizing the accuracy and maintaining the flexibility across a variety of reasoning tasks.

\subsection{Impact of Pool Size on CDW-CoT}

Similar to the sensitivity analysis of temperature effects, we make the analysis to explore the impact of Pool Size $S$ on the CDW-CoT framework. This parameter is important as it controls the number of prompt candidates extracted from clusters.

We varied the pool size from 10 to 40 to assess how the quantity of available prompt candidates impacts the model's performance. This investigation was conducted using the LLaMA2(13B) model on two datasets: CommonsenseQA and MultiArith.

\begin{figure}[htb]
    \centering
    \includegraphics[width=\columnwidth]{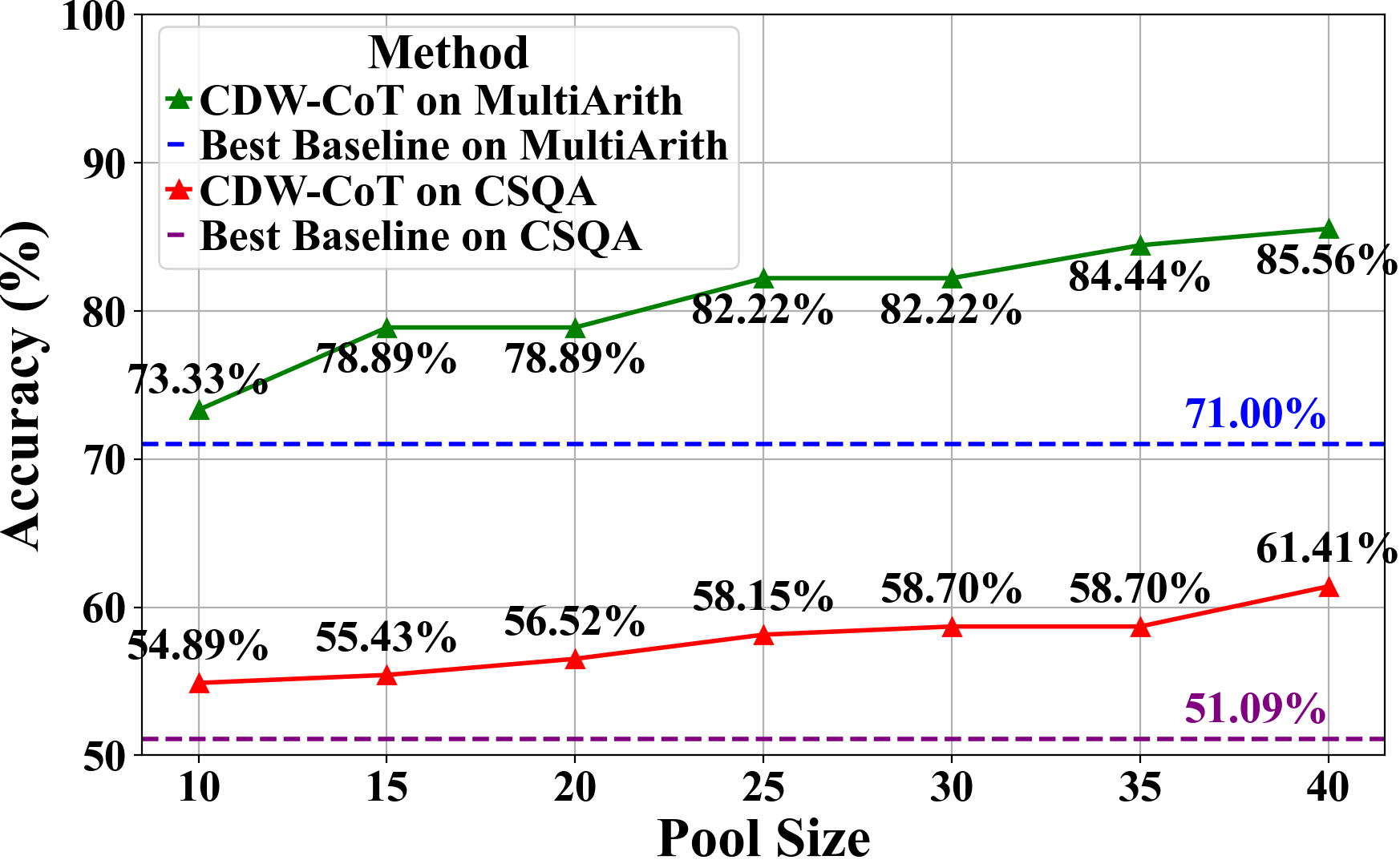} 
    \caption{Impact of pool size on CDW-CoT on CommonsenseQA and MultiArith datasets.}
    \label{fig:pool_size_effect}
\end{figure}
Based on the analysis and trends from Figure~\ref{fig:pool_size_effect}, we observe that increasing the pool size consistently enhances the model’s performance across both datasets. As expected, a larger candidate pool allows the CDW-CoT framework to better explore diverse reasoning paths. For MultiArith, accuracy steadily improves from 73.33\% at a pool size of 10 to 85.56\% at a pool size of 40. Similarly, for CommonsenseQA, accuracy increases from 54.89\% to 61.41\% as the pool size grows. 

While a larger pool increases reasoning diversity and improves the accuracy, it also increases the computational costs. In our experiments, we chose a pool size of 40 as an optimal balance between performance gains and efficiency. This selection ensures that the CDW-CoT framework achieves high accuracy across different reasoning tasks without incurring excessive computational overhead, effectively balancing decision quality and resource use.

\section{Conclusion}
In this paper, we propose a novel CoT method named CDW-CoT to enhance the adaptability and accuracy of LLMs in complex reasoning tasks. Our method introduces the clustering to categorize the datasets into tailored prompt pools, improving the representative ability to diverse data characteristics. It calculates an optimal prompt probability distribution for each cluster, enabling targeted reasoning that aligns with its unique characteristics. By designing the distance-weighted prompt selection, CDW-CoT dynamically adjusts the reasoning strategies based on the proximity to cluster centers, demonstrating superior performance over traditional methods across six datasets. Future work includes reducing computational overhead and extending applicability to multimodal tasks like image-text reasoning.

\section{Acknowledgements}
This work is supported in part by the National Natural Science Foundation of China (No. 62276079), Young Teacher Development Fund of Harbin Institute of Technology IDGA10002071, Research and Innovation Foundation of Harbin Institute of Technology IDGAZMZ00210325 and the Special Funding Program of Shandong Taishan Scholars Project.

\bibliography{aaai25}

\end{document}